\documentclass[conference]{IEEEtran}
\IEEEoverridecommandlockouts
\usepackage{cite}
\usepackage{amsmath,amssymb,amsfonts}
\usepackage{algorithmic}
\usepackage{algorithm}
\usepackage{graphicx}
\usepackage{textcomp}
\usepackage{xcolor}

\DeclareMathOperator*{\argmax}{arg\,max}

\def\BibTeX{{\rm B\kern-.05em{\sc i\kern-.025em b}\kern-.08em
    T\kern-.1667em\lower.7ex\hbox{E}\kern-.125emX}}
\begin{document}

\title{Active Hierarchical Imitation and Reinforcement Learning\\
{\footnotesize \textsuperscript{}}
\thanks{}
}

\author{\IEEEauthorblockN{Yaru Niu}
\IEEEauthorblockA{\textit{School of Electrical and Computer Engineering} \\
\textit{Georgia Institute of Technology}\\
yaruniu@gatech.edu}
\and
\IEEEauthorblockN{Yijun Gu}
\IEEEauthorblockA{\textit{School of Computer Science} \\
\textit{Georgia Institute of Technology}\\
yjgu@gatech.edu}
}

\maketitle

\begin{abstract}
Humans can leverage hierarchical structures to split a task into sub-tasks and solve problems efficiently. Both imitation and reinforcement learning or a combination of them with hierarchical structures have been proven to be an efficient way for robots to learn complex tasks with sparse rewards. However, in the previous work of hierarchical imitation and reinforcement learning, the tested environments are in relatively simple 2D games, and the action spaces are discrete. Furthermore, many imitation learning works focusing on improving the policies learned from the expert polices that are hard-coded or trained by reinforcement learning algorithms, rather than human experts. In the scenarios of human-robot interaction, humans can be required to provide demonstrations to teach the robot, so it is crucial to improve the learning efficiency to reduce expert efforts, and know human's perception about the learning/training process. In this project, we explored different imitation learning algorithms and designed active learning algorithms upon the hierarchical imitation and reinforcement learning framework we have developed. We performed an experiment where five participants were asked to guide a randomly initialized agent to a random goal in a maze. Our experimental results showed that using DAgger and reward-based active learning method can achieve better performance while saving more human efforts physically and mentally during the training process.
\end{abstract}


\section{Introduction}
Hierarchical structures can help human to split a task into sub-tasks and solve problems efficiently. Both imitation and reinforcement learning or a combination of them with hierarchical structures have been proven to be an efficient way for robots to learn complex tasks ~\cite{levy2018hac, le2018hirl}.

Imitation Learning (IL) is one of the efficient algorithms we use with an oracle or human demonstrators with (near) optimal performance. Sun and Bagnell compared the theoretical bound of performance using Reinforcement Learning (RL) with the bound using IL under some simple cases, and showed IL outperforms RL for a polynomial factor in general Markov Decision Process (MDP)~\cite{sun2017deeply}. Hierarchical Reinforcement Learning (HRL) is another efficient algorithm which uses temporal abstractions to help the agent to explore in more semantically meaningful action space, and thus improve the sample efficiency of RL algorithm~\cite{nachum2019does}. Does combining two approaches further help us in complex tasks? In Hierarchical Imitation and Reinforcement Learning (HIRL)~\cite{le2018hirl}, the authors combined both methods in discrete state-action space, together with some strategies that restrict the place where learning occurs. They showed that their approach could decrease the expert's cost in training the agent, therefore, outperform other hierarchical approaches. Therefore, we developed a similar approach in this work as in HIRL: training high-level controller with IL and low-level controller with RL. Our task was built on a maze-navigation domain in continuous state and action space where we did not directly apply HIRL.

In the continuous setting, we seek to learn a set of policies together that is parameterized by the goal, and the problem is often referred to as contextual policy learning or multi-task learning. Some typical examples are locomotion task and robot arm manipulation task like\cite{peng2017deeploco,andrychowicz2017hindsight}. In our task, both the high-level controller and low-level controller are parameterized by their goal. Training such a policy is time-consuming. The model training in ~\cite{peng2017deeploco} requires several days of computation. As mentioned above, we expect that the agent can learn fast if it combines IL and uses a hierarchical structure. To further increase sample efficiency in this setting, we develop active learning methods and compare them with directly employing imitation learning on the high-level controller.  

Furthermore, to evaluate how our proposed methods perform in a real human-robot interaction scenario and investigate how human feel about the learning/training process, we performed an experiment where five participants were asked to guide a randomly initialized agent to a random target in a maze. The results showed that using our hierarchical imitation and reinforcement learning framework with reward-based active learning method can achieve better performance than others while saving more human efforts physically and mentally during the training process.

\section{Related Work}
\subsection{Interactive Imitation learning}
Sequential prediction problems arise commonly in practice, and interactive imitation learning incorporates the learning processes in the teaching phase. Ross and Bagnell proposed an iterative meta-algorithm DAgger~\cite{ross2011dagger}, which trains a stationary deterministic policy, that can be seen as a no-regret algorithm in an online learning setting. The approach is guaranteed to perform well under its induced distribution of states. However, in many cases the learner can only get access to a sub-optimal expert policy. Sun et al. proposed an interactive imitation learning algorithm, AggreVaTeD, on sequential prediction tasks~\cite{sun2017deeply}. AggreVaTeD can reach expert-level and even super-expert performance. We integrated the interactive imitation learning algorithms in our hierarchical framework to enable the human to teach the agent high-level policy.

\subsection{Hierarchical Reinforcement Learning}
Building agents that can learn hierarchical policies is a longstanding problem in Reinforcement Learning, for example, decompose MDP into smaller MDPs~\cite{dietterich2000hierarchical} or feudal RL that introduces spacial or temporal abstraction into the learning process~\cite{dayan1993feudal}. Some theoretical results are given, where we can adopt options into MDPs and then obtain a semi-Markov decision process for high-level policies so that the agent can learn faster~\cite{sutton1999between}. Recently, there are many new pieces of research that incorporates deep neural networks with hierarchical reinforcement learning. There are several automated HRL techniques that can work in discrete domains with a manually designed intrinsic reward for low-level policy~\cite{kulkarni2016hierarchical}. Further works like Option-Critic architecture~\cite{bacon2017option} and FeUdal Network~\cite{vezhnevets2017feudal} address the problem of predefined subgoals as well as extend HRL to continuous space. The main problem of HRL is that high-level policy learning is unstable due to the fact that low-level policy is changing. To address this issue, we use Hindsight Experience Replay~\cite{andrychowicz2017hindsight} which was proposed in Hierarchical Actor-Critic (HAC)~\cite{levy2018hac} or posterior probability optimization of the subgoal given low-level trajectories~\cite{nachum2018hiro}. While most of the HRL methods only use a two-layer structure, HAC extends the framework to an arbitrary number of layers and each layer can be trained in parallel. It is further shown that HRL can be regarded as a multi-agent RL task and the non-stationary problem can be addressed in this view~\cite{kreidieh2019inter}.

\subsection{Active Learning}
Active Learning is a popular approach for classification in the semi-supervised learning setting. The expert provides labels only when the agent asks for labels. The approach aims at saving the expert's labeling cost and it is very similar to our goal. In the context of LfD, researchers have proposed active learning methods based on confidence calculation~\cite{li2006confidence, chernova2009interactive} or even learn how to active learn~\cite{bullard2019active}. Despite the fact that it is a popular LfD method, we can only find a few pieces of research on the combination of multi-task learning and active learning~\cite{fabisch2014active,silver2012active}. The method we adopted in our task is inspired by the work~\cite{silver2012active}, where the author uses active learning for navigation tasks and the agent is selecting tasks (starting and ending points) for the human demonstrator. We also adopted the method by Hafner~\cite{hafner2018reliable}, where they use noise contrastive priors to estimate the reliable uncertainty of the neural networks.

\subsection{Combining RL and LfD}
The idea of combining IL and RL is not new ~\cite{nair2018overcoming,hester2018deep}. The proposed method Deep Q-Learning from Demonstration (DQfD) is taking IL as a "pre-training" step by pre-populating the replay buffer with demonstrations~\cite{hester2018deep}. The pre-populated training examples can be used to warm start the agent. In HIRL, the combination of IL and RL is in the form of interaction between meta controller and low-level controller instead of warm start the agent. However, previous works only focused on the application in discrete state and action spaces~\cite{nair2018overcoming,hester2018deep,le2018hirl}. In contrast, we built our pre-training IL and HRL in continuous tasks with continuous feedback. 

\section{Methods}
\subsection{Interactive Imitation Learning}
In our method, we tried to impelement both DAgger and AggreVaTeD as our interactive imitation learning methods. We found that DAgger can fit in our hierarchical imitation and reinforcement learning framework well and learn the high-level policy quickly within reasonable number of interactive training episodes, while AggreVaTeD cannot improve the policy with the same scale of episodes. We thought that AggreVaTeD fails in our setting because AggreVaTeD performs policy gradient to update the learner's policy using the reward information in the environment, rather than directly learn from expert demonstrations like DAgger. In our settings, the rewards from the environment are sparse, and human can only afford reasonable number of interactive training time, so the policy learned from AggreVaTeD is very likely to gain nothing during this period. In this case, we finally used DAgger as our high-level policy learning method for the agent, and the following discussions are all based on DAgger.

\subsection{Hierarchical Imitation and Reinforcement Learning}
\begin{algorithm}[h]
\caption{Hierarchical Imitation and Reinforcement Learning (High-Level)}
\label{alg:HHIRL}
\begin{algorithmic}[1]
 \renewcommand{\algorithmicrequire}{\textbf{Input:}}
 \renewcommand{\algorithmicensure}{\textbf{Output:}}
\REQUIRE Environment, and pretrained low-level policy
\ENSURE Learned high-level policy
\STATE Randomly initialize high-level policy $\pi_{\text{HI}}$
\STATE Initialize dataset $\mathcal{D} \leftarrow \emptyset$
\STATE Initialize mixing factor $\beta \leftarrow{} 1.0$
\REPEAT
    \STATE Sample end goal $g$ and initial state $s_0$ from environment
    \REPEAT
        \STATE Get $(s,\pi^*(s))$ from demonstrator 
        \STATE Generate subgoal $g_{\text{LO}} \leftarrow \beta\pi_{\text{HI}} + (1-\beta)\pi^*$
        \STATE $\mathcal{D} \leftarrow \mathcal{D} \bigcup \{(s,a)\}$
        \STATE Run low-level policy for fix horizon $H$
        \IF{Reach update episodes}
            \REPEAT
            \STATE $\pi_{\text{HI}} \leftarrow \pi_{\text{HI}} - \nabla \ell(s,a)$
                \UNTIL{number of update}
            \STATE $\beta \leftarrow \beta / d$
        \ENDIF
        \STATE Execute low-level policy for horizon $H$ given $g_{\text{LO}}$
    \UNTIL{Goal reached or maximum number of action taken}
\UNTIL{100 episodes}
\end{algorithmic}
\end{algorithm}

Here we introduce the Hierarchical Imitation Reinforcement Learning (HIRL) framework that can effectively learn multi-level policies in continuous spaces. We constructed the environment with a two-level hierarchy; the high level corresponds to choosing subtasks to reach the final goal and the low level corresponds to executing subtasks. We trained our high-level meta-controller using DAgger and low-level controller using a revised Deep Deterministic Policy Gradient (DDPG) method ~\cite{lillicrap2015continuous}. The algorithm of HIRL is shown in Alg.~\ref{alg:HHIRL}.


\subsection{Active Learning}
To improve the learning performance and efficiency of HIRL, and make it use-friendly during human-robot interaction in a goal-achieving task, we proposed the following active learning methods.

\subsubsection{Noise-Based}
In a goal-achieving task, the agent state and goal state might not have the same dimension, and we assume that the agent state dimension includes the goal state dimension. And the noise-based method is based on the assumption that adding noise to the sub-dimension of the agent state should not change the expect of the optimal action, or it can be meaningless to calculate the variance of the output.

Let $s$ represent the full dimension state of the agent state, $s_g$ the sub-dimension of the agent state which is consistent with the goal state. First, we randomly choose a number of $s_g$ as the candidates of the initialized sub-state. Then we add random uniform noise to other sub-dimension states $s_o$ of the agent state rather than $s_g$. Then we get $n$ agent state regarding the same $s_g$. Then we choose the $s_g$ based on the equation written as 
\begin{equation}
    s_g^* = \argmax_{s_g \in S_g} \sum_{i=1}^n[\pi_{\theta}(s_i(s_g)) - \bar{\pi}_{\theta_i}(s_i(s_g))]^2   \label{eq_noise}
\end{equation} 
where $S_g$ is the candidate pool of the $s_g$, and $s_i(\cdot)$ is the function to get the state after adding an random uniform noise to $s_o$, and $\pi_{\theta}$ is the current policy. In our task, $s_g$ is the cartesian position of the agent. The algorithm is shown as Alg.~\ref{alg:AL_noise}.

\begin{algorithm}[h]
\caption{Noise-Based Active Learning}
\label{alg:AL_noise}
\begin{algorithmic}[3]
 \renewcommand{\algorithmicrequire}{\textbf{Input:}}
 \renewcommand{\algorithmicensure}{\textbf{Output:}}
\REQUIRE Environment, candidate pool size, and number of different noises, current high-level policy $\pi_{\theta}$, end goal $g$
\ENSURE Initialized position $s_g^*$
\STATE Initialize noise buffer $\mathcal{N} \leftarrow \emptyset$
\STATE Initialize state dataset $\mathcal{S} \leftarrow \emptyset$
\REPEAT
    \STATE Randomize noise $n$
    \STATE $\mathcal{N} \leftarrow \mathcal{N} \bigcup \{n\}$
\UNTIL number of different noises 
\REPEAT
    \STATE Randomly choose a position $s_g$ from environment
    \REPEAT 
        \STATE Add noise from $\mathcal{N}$ to $s_o$
        \STATE Get new state $s$
        \STATE $\mathcal{S} \leftarrow \mathcal{S} \bigcup \{s\}$
    \UNTIL number of different noises
    \STATE Compute the variance with $S$, $g$ and $\pi_{\theta}$ based on equation (\ref{eq_noise})
    \STATE Store variance
    \STATE $\mathcal{S} \leftarrow \emptyset$
\UNTIL{candidate pool size}
\STATE Choose $s_g^*$ based on equation (\ref{eq_noise})
\end{algorithmic}
\end{algorithm}

\subsubsection{Multi-Policy}
Another AL method we have employed is similar to bagging in ensemble learning. We trained $n$ policies with the same training dataset and tested how much these policies agree given a randomly selected state. This method can be formally written as
\begin{equation}
    s_g^* = \argmax_{s_g \in S_g} \sum_{i=1}^n[\pi_{\theta_i}(s(s_g)) - \bar{\pi}_{\theta_i}(s(s_g))]^2   \label{eq_bagging}
\end{equation}
where $S_g$ is the candidate pool of the $s_g$, and $s(\cdot)$ is a fixed mapping function from $s_g$ to , and $\pi_{theta_i}$ is train $i$ policy. The algorithm is shown in Alg. ~\ref{alg:AL_bagging}.

\begin{algorithm}[H]
\caption{Multi-Policy Active Learning}
\label{alg:AL_bagging}
\begin{algorithmic}[3]
 \renewcommand{\algorithmicrequire}{\textbf{Input:}}
 \renewcommand{\algorithmicensure}{\textbf{Output:}}
\REQUIRE Environment, candidate pool size, and number of different noises, current high-level policy bag $\{\pi_{\theta_i}\}_{i=1}^{n}$, end goal $g$
\ENSURE Initialized position $s_g^*$
\REPEAT
    \STATE Randomly choose a position $s_g$ from environment
    \STATE Get new state $s$
    \STATE Compute the variance with $s$, $g$ and $\{\pi_{\theta_i}\}_{i=1}^{n}$ based on equation (\ref{eq_bagging})
    \STATE Store variance
\UNTIL{candidate pool size}
\STATE Choose $s_g^*$ based on equation (\ref{eq_bagging})
\end{algorithmic}
\end{algorithm}

\subsubsection{Reward-Based}
Different from the previous two methods that actively initialize the position (sub-state) of the agent at the start of each episode, reward-based active learning utilize the reward information from the environment and choose more useful data to store in the experience replay buffer. We assume the agent will only get the reward when it reaches the final goal, and each episode will end when the agent reach the final goal or the time step achieves its maximum. Within each episode, if the agent reaches the final goal, the data aggregation will follow the same mechanism as described in Alg. ~\ref{alg:HHIRL}. If the agent cannot reach the goal, the data in $\mathcal{D}$ will be popped from the experience replay buffer following a probability:
\begin{equation}
p = e^{-t_d}
\end{equation}
where $t_d$ is a time step distance to the end of the episode. The algorithm is shown in Alg. ~\ref{alg:AL_reward}.

\begin{algorithm}[H]
\caption{Reward-Based Active Learning}
\label{alg:AL_reward}
\begin{algorithmic}[3]
 \renewcommand{\algorithmicrequire}{\textbf{Input:}}
 \renewcommand{\algorithmicensure}{\textbf{Output:}}
 \renewcommand{\algorithmicrequire}{\textbf{Input:}}
 \renewcommand{\algorithmicensure}{\textbf{Output:}}
\STATE Intialize $t \leftarrow 0$
\FOR{each episode}
    \STATE $t \leftarrow t+1$
    \STATE Sample end goal $g$ and initial state $s_0$ from environment
    \REPEAT
        \STATE Get $(s^t,\pi^*(s^t))$ from demonstrator 
        \STATE Generate subgoal $g_{\text{LO}} \leftarrow \beta\pi_{\text{HI}} + (1-\beta)\pi^*$
        \STATE $\mathcal{D} \leftarrow \mathcal{D} \bigcup \{(s^t,a^t)\}$
        \STATE Run low-level policy for fix horizon $H$
        \STATE Update high-level policy: $\pi_{\text{HI}} \leftarrow \pi_{\text{HI}} - \nabla \ell(s^t,a^t)$
        \STATE Update $\beta$: $\beta \leftarrow \beta / d$
        \STATE Execute low-level policy for horizon $H$ given $g_{\text{LO}}$
    \UNTIL{Goal reached or maximum number of action taken}
    \IF{Goal is nor reached}
        \FOR{$t_d = 0$ to length of the previous episode}
        \STATE With probability $e^{-t_d}$, $\mathcal{D}$ pops $(s^{t-t_d},a^{t-t_d})$
        \ENDFOR
    \ENDIF
\ENDFOR
\end{algorithmic}
\end{algorithm}

\begin{figure}
\centerline{\includegraphics[width=70mm, scale=0.5]{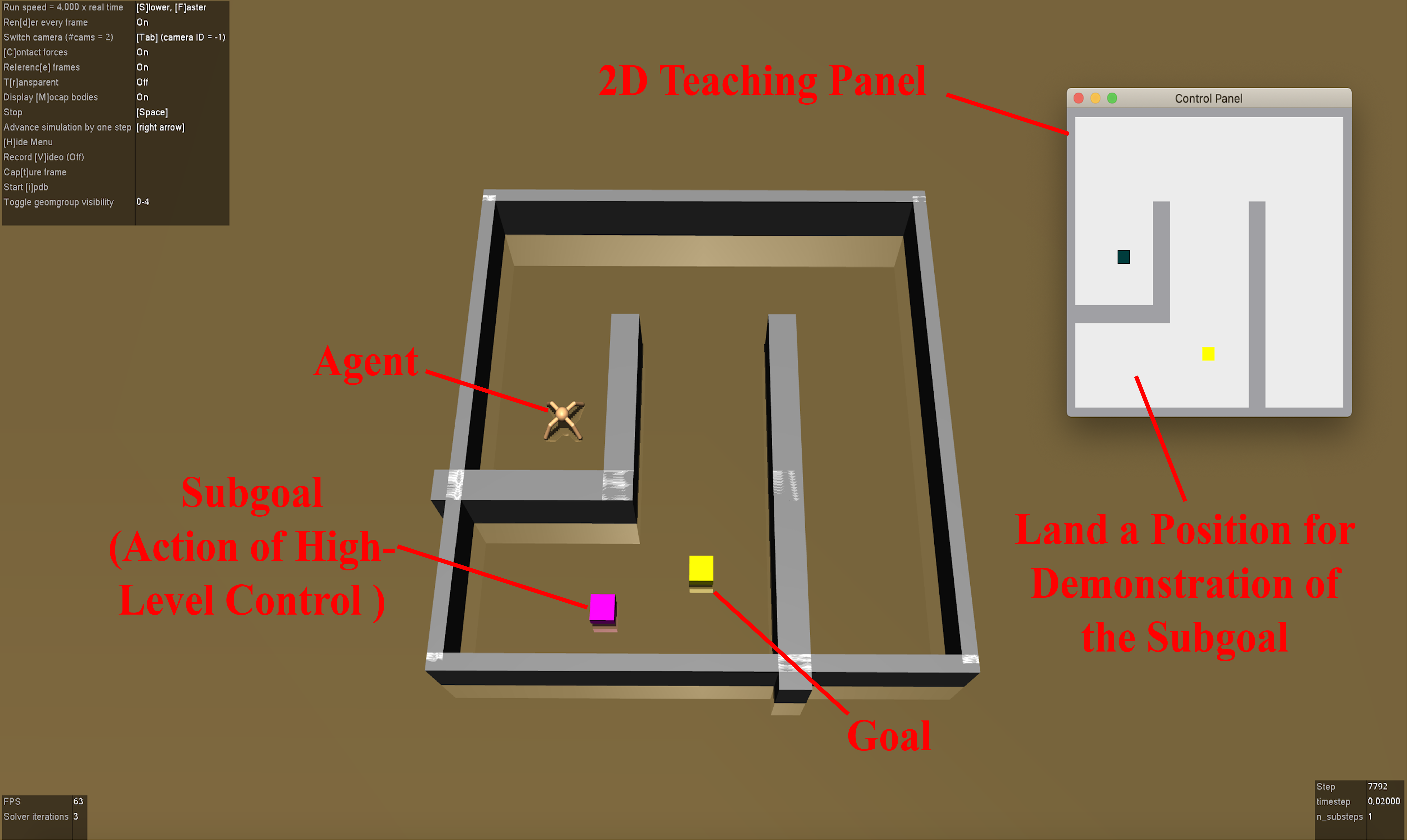}}
\caption{The task overview of our experiment.}
\label{fig:task_overview}
\end{figure}

\begin{figure*}
\centering
\includegraphics[width=0.30\textwidth, scale=0.6]{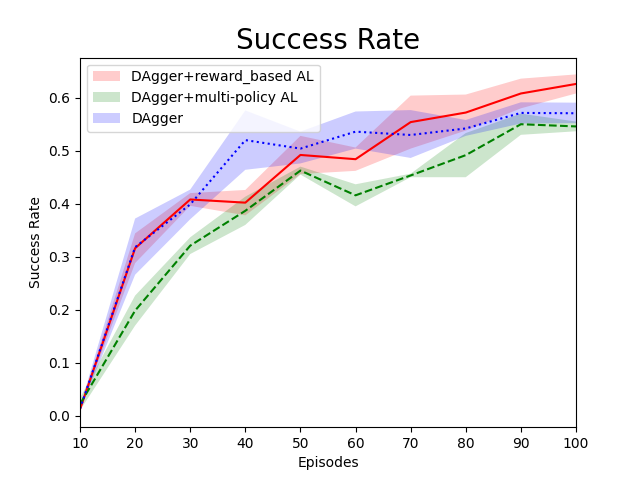}
\includegraphics[width=0.30\textwidth, scale=0.6]{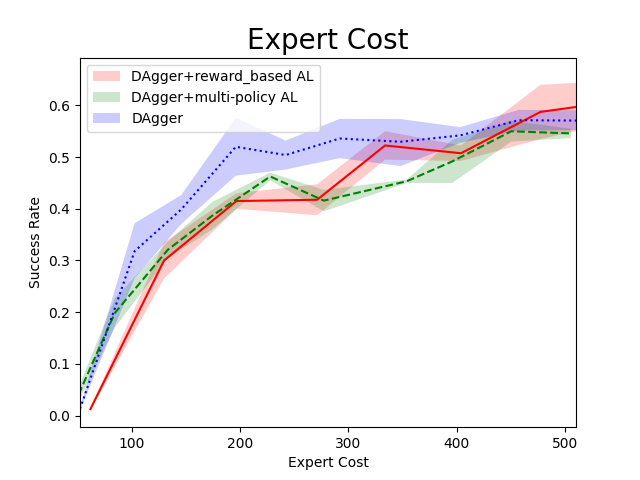}
\includegraphics[width=0.30\textwidth, scale=0.6]{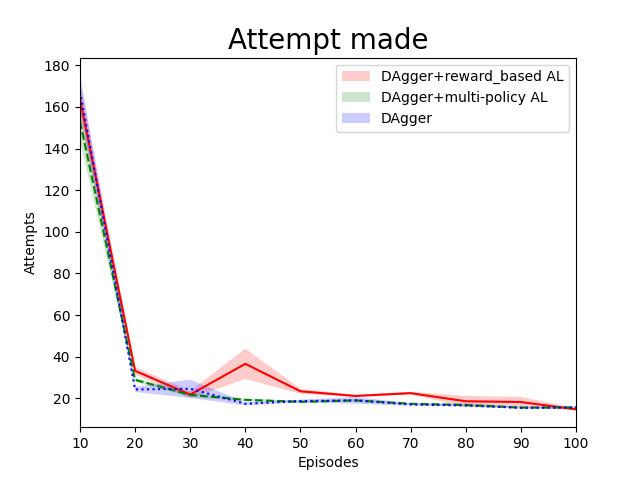}
\caption{\label{fig:3results}Experiment Results over 300 test episodes. (a) Success rate per episode. (b) Success rate versus the expert cost. (c) Attempts per episode.} 
\end{figure*}

\begin{figure}
\centerline{\includegraphics[width=0.45\textwidth, scale=0.5]{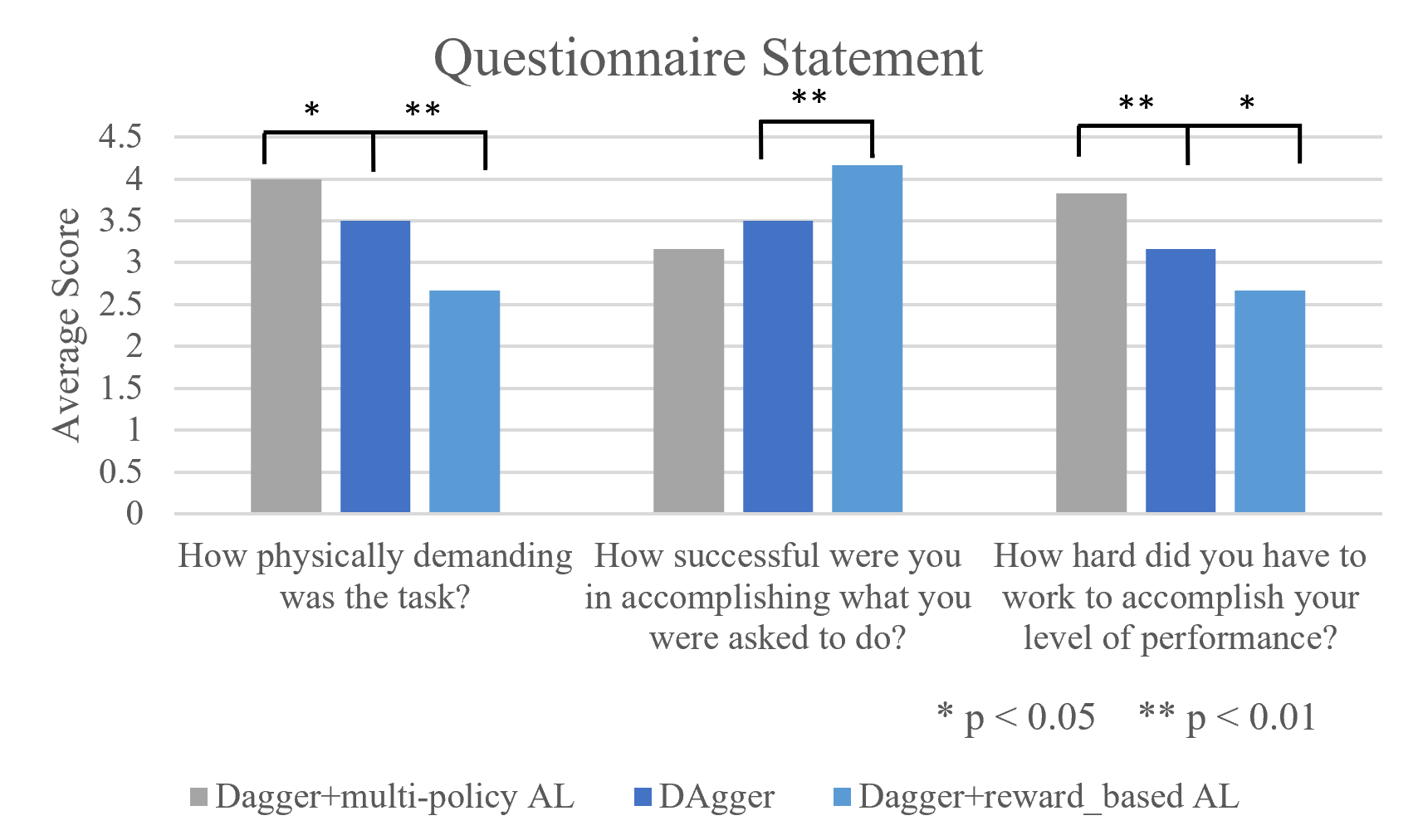}}
\vspace{-0.2cm}
\caption{\label{fig:likert}Comparison of 5-point scale questionnaire responses for \emph{DAgger}, \emph{DAgger+Multi-policy AL}, and \emph{DAgger+Reward-based AL}. Responses are averaged over all participants. p-values are computed with one-way ANOVA test.}
\vspace{-0.5cm}
\end{figure}

\section{Human user study}
We evaluated our HIRL framework using Multi-Policy active learning (\emph{DAgger+Multi-Policy AL}), and using Reward-Based active learning (\emph{DAgger+Reward-Based AL}), compared to its vanilla version (\emph{DAgger}). We conducted a within-subjects user study with five participants. As Noise-based active learning method has similar results with Multi-Policy one, we only evaluated the latter one in the experiment.  

\subsection{Task Environment}
We built our experiment on a 3D maze navigation domain developed in MuJoCo \cite{todorov2012mujoco} as shown in Fig.~\ref{fig:task_overview}. In each episode, the agent (standard Rllab Ant with a gear range of (-30, 30)) needs to navigate from one corner of the maze with randomly generated layouts to a target marked in yellow with a time limit of 500 steps. Grey rectangles are obstacles (wall), which the agent needs to avoid for survival. The contextual in-formation the agent receives is the pixel representation of a bird’s-eye view of the environment, including the positions of the agent and the final goal.

\subsection{Experimental Procedure}
We recruited 5 able-bodied human participants (3 females and 2 males). We recruited participants to meet the following criteria: 8 years of age or older; able-bodied; no cognitive or visual impairments; fluent in spoken and written English; and not diagnosed with epilepsy. Participant ages ranged from 23 to 25 years old. For each participant, we conducted a total of 3 trials for the three algorithms: \emph{DAgger}, \emph{DAgger+Multi Policy AL}, and \emph{DAgger+Reward-based AL}. Each trial consists of 100 episodes. We randomized the ordering of the three methods for each participant according to randomized counter balancing design.

During the study, the participants moved the mouse the provide demonstrations of the subgoal positions to help agents to navigate to the target. The participants were instructed to show a subgoal position towards the target but also make sure not to be too far from the current position of the agent. Prior to the study, we gave participants an unscripted practice trial with 30 episodes to familiarize themselves on interacting with the agent and how to provide robust demonstrations to accomplish the tasks.

\subsection{Objective Measures}
We used the following measures as objective measures of performance:
    \begin{itemize}
        \item \textbf{Success rate:} average task success percentages over 100 test episodes; task success is defined by the agent reaching the final goal.
        \item \textbf{Expert cost:} average number of demonstrations in one episode over 100 test episodes.
        \item \textbf{Attempts:} average number of subgoals generated to reach the final goal in one episode over 100 test episodes.
    \end{itemize}
\subsection{Subjective Measures}
Seeking to capture the participants' perceptions of the policy learning abilities, we used the questionnaire with three statements pertaining to perceptions of the task physical demand, performance, and human effort. For each statement, participants were asked to rate how much they agreed with the statement on the interval scale from 1 to 5. We based this 5-Point Likert Scale questionnaire on NASA-TLX~\cite{hart2006nasa}. \\
The four statements follow:
\begin{itemize}
    \item How physically demanding was the task? (1 = Very low, 5 = Very high)
    \item How successful were you in accomplishing what you were asked to do? (1 = Failure, 5 = Perfect)
    \item How successful were you in accomplishing what you were asked to do? (1 = Very low, 5 = Very high)
\end{itemize}


\section{Results}
\subsection{Objective Measures}
We compared our AL-based HIRL algorithms with the vanilla version. Fig.~\ref{fig:3results} displays the results of the comparison.  
\subsubsection{Success rate} Fig.~\ref{fig:3results} (left) shows the success rate of three algorithms over 100 episodes. \emph{DAgger+Reward-based AL} outperforms \emph{DAgger} and achieves consistently highest success rate among the three algorithms after about 60 episodes. It exceeds 60\% in fewer than 100 episodes. \emph{DAgger+Multi Policy AL} performs worse than \emph{DAgger} and has largest variation among the three algorithms. The figure displays the median as well as the range from minimum to maximum success rate over 5 random executions of the algorithms.
\subsubsection{Expert cost} Fig.~\ref{fig:3results} (middle) shows the same success rate, but as a function of the expert cost. \emph{DAgger} employs less expert demonstrations compared to other two algorithms when achieving a success rate lower than 50\%. To achieve a success rate more than 50\%, \emph{DAgger+Reward-based AL} achieves most savings in expert cost.  \emph{DAgger+Multi Policy AL} needs the most expert costs within 100 episodes.
\subsubsection{Attempts} Fig.~\ref{fig:3results} (right) shows that all of the three algorithms decrease suddenly when trained after the first 20 episodes and have a slow decrease after 20 episodes. There is no obvious difference among them.

Through the results, we concluded that \emph{DAgger+Reward-based AL} performs better than the flat version (\emph{DAgger}) while \emph{DAgger+Multi-policy AL} performs worse than \emph{DAgger}.

\subsection{Subjective Measures}
Fig.~\ref{fig:likert} displays a comparison of questionnaire responses between AL-based HIRL algorithms and vernilla version when responses are averaged over all participants for each questionnaire statement. To determine the statistical difference between algorithms, we applied a one-way ANOVA test. The computed p-values are depicted in Fig.~\ref{fig:likert}. 
\begin{itemize}
    \item \textbf{\emph{DAgger} and \emph{DAgger+Multi-Policy AL}: } Participants provided higher responses related to physical demand and effort and lower response related to success in which robot used \emph{DAgger+Multi-Policy AL} algorithm to learn from demonstrations. From Fig.~\ref{fig:likert}, we observed a statistically significant difference at the $p<.05$ level for questions relating to physical demand, and a less significant difference at the $p<.001$ level for effort question between \emph{DAgger} and \emph{DAgger+Multi-Policy AL}. 
    \item \textbf{\emph{DAgger} and \emph{DAgger+Reward-based AL}: } Participants provided lower responses related to physical demand and effort and higher response related to success in which robot used \emph{DAgger+Reward-based AL} algorithm to learn from demonstrations. From Fig.~\ref{fig:likert}, we observed a statistically significant difference at the $p<.001$ level for questions relating to physical demand, task success and a less significant difference at the $p<.05$ level for effort between \emph{DAgger} and \emph{DAgger+Reward-based AL}.
\end{itemize}

These results indicate that participants perceived a noticeable improvement in an agent's performance and easiness when using \emph{DAgger+Reward-based AL} to provide guided demonstrations while perceive a decrease when using \emph{DAgger+Multi-Policy AL}.

\section{Conclusion and Future Work}
We explored active learning algorithms with hierarchical imitation and reinforcement learning framework. We presented two active learning-based HIRL algorithms: \emph{DAgger+Multi-policy AL} and \emph{DAgger+Reward-based AL}. We also provided evidence that \emph{DAgger+Reward-based AL} can improve the performance and reduce the cost of expert feedback in hierarchical imitation and reinforcement learning. Overall, our results suggest that active reward learning can be used to help speed up learning from demonstration feedback and can be extended in the future work. 

\section*{Acknowledgment}
This is a course project for CS 7633 Human Robot Interaction in Georgia Institute of Technology. We thank Prof. Sonia Chernova for feedback about this work and Zuoxing Tang for assistance. 

\vspace{12pt}
\bibliographystyle{plain}
\bibliography{references}

\begin{thebibliography}{10}

\bibitem{andrychowicz2017hindsight}
Marcin Andrychowicz, Filip Wolski, Alex Ray, Jonas Schneider, Rachel Fong,
  Peter Welinder, Bob McGrew, Josh Tobin, OpenAI~Pieter Abbeel, and Wojciech
  Zaremba.
\newblock Hindsight experience replay.
\newblock In {\em Advances in Neural Information Processing Systems}, pages
  5048--5058, 2017.

\bibitem{bacon2017option}
Pierre-Luc Bacon, Jean Harb, and Doina Precup.
\newblock The option-critic architecture.
\newblock In {\em Thirty-First AAAI Conference on Artificial Intelligence},
  2017.

\bibitem{bullard2019active}
Kalesha Bullard, Yannick Schroecker, and Sonia Chernova.
\newblock Active learning within constrained environments through imitation of
  an expert questioner.
\newblock {\em arXiv preprint arXiv:1907.00921}, 2019.

\bibitem{chernova2009interactive}
Sonia Chernova and Manuela Veloso.
\newblock Interactive policy learning through confidence-based autonomy.
\newblock {\em Journal of Artificial Intelligence Research}, 34:1--25, 2009.

\bibitem{dayan1993feudal}
Peter Dayan and Geoffrey~E Hinton.
\newblock Feudal reinforcement learning.
\newblock In {\em Advances in neural information processing systems}, pages
  271--278, 1993.

\bibitem{dietterich2000hierarchical}
Thomas~G Dietterich.
\newblock Hierarchical reinforcement learning with the maxq value function
  decomposition.
\newblock {\em Journal of artificial intelligence research}, 13:227--303, 2000.

\bibitem{fabisch2014active}
Alexander Fabisch and Jan~Hendrik Metzen.
\newblock Active contextual policy search.
\newblock {\em The Journal of Machine Learning Research}, 15(1):3371--3399,
  2014.

\bibitem{hafner2018reliable}
Danijar Hafner, Dustin Tran, Alex Irpan, Timothy Lillicrap, and James Davidson.
\newblock Reliable uncertainty estimates in deep neural networks using noise
  contrastive priors.
\newblock {\em arXiv preprint arXiv:1807.09289}, 2018.

\bibitem{hart2006nasa}
Sandra~G Hart.
\newblock Nasa-task load index (nasa-tlx); 20 years later.
\newblock In {\em Proceedings of the human factors and ergonomics society
  annual meeting}, volume~50, pages 904--908. Sage publications Sage CA: Los
  Angeles, CA, 2006.

\bibitem{hester2018deep}
Todd Hester, Matej Vecerik, Olivier Pietquin, Marc Lanctot, Tom Schaul, Bilal
  Piot, Dan Horgan, John Quan, Andrew Sendonaris, Ian Osband, et~al.
\newblock Deep q-learning from demonstrations.
\newblock In {\em Thirty-Second AAAI Conference on Artificial Intelligence},
  2018.

\bibitem{kreidieh2019inter}
Abdul~Rahman Kreidieh, Samyak Parajuli, Nathan Lichtle, Yiling You, Rayyan
  Nasr, and Alexandre~M Bayen.
\newblock Inter-level cooperation in hierarchical reinforcement learning.
\newblock {\em arXiv preprint arXiv:1912.02368}, 2019.

\bibitem{kulkarni2016hierarchical}
Tejas~D Kulkarni, Karthik Narasimhan, Ardavan Saeedi, and Josh Tenenbaum.
\newblock Hierarchical deep reinforcement learning: Integrating temporal
  abstraction and intrinsic motivation.
\newblock In {\em Advances in neural information processing systems}, pages
  3675--3683, 2016.

\bibitem{le2018hirl}
Hoang~M Le, Nan Jiang, Alekh Agarwal, Miroslav Dud{\'\i}k, Yisong Yue, and Hal
  Daum{\'e}~III.
\newblock Hierarchical imitation and reinforcement learning.
\newblock {\em arXiv preprint arXiv:1803.00590}, 2018.

\bibitem{levy2018hac}
Andrew Levy, George Konidaris, Robert Platt, and Kate Saenko.
\newblock Learning multi-level hierarchies with hindsight.
\newblock 2018.

\bibitem{li2006confidence}
Mingkun Li and Ishwar~K Sethi.
\newblock Confidence-based active learning.
\newblock {\em IEEE transactions on pattern analysis and machine intelligence},
  28(8):1251--1261, 2006.

\bibitem{lillicrap2015continuous}
Timothy~P Lillicrap, Jonathan~J Hunt, Alexander Pritzel, Nicolas Heess, Tom
  Erez, Yuval Tassa, David Silver, and Daan Wierstra.
\newblock Continuous control with deep reinforcement learning.
\newblock {\em arXiv preprint arXiv:1509.02971}, 2015.

\bibitem{nachum2018hiro}
Ofir Nachum, Shixiang~Shane Gu, Honglak Lee, and Sergey Levine.
\newblock Data-efficient hierarchical reinforcement learning.
\newblock In {\em Advances in Neural Information Processing Systems}, pages
  3303--3313, 2018.

\bibitem{nachum2019does}
Ofir Nachum, Haoran Tang, Xingyu Lu, Shixiang Gu, Honglak Lee, and Sergey
  Levine.
\newblock Why does hierarchy (sometimes) work so well in reinforcement
  learning?
\newblock {\em arXiv preprint arXiv:1909.10618}, 2019.

\bibitem{nair2018overcoming}
Ashvin Nair, Bob McGrew, Marcin Andrychowicz, Wojciech Zaremba, and Pieter
  Abbeel.
\newblock Overcoming exploration in reinforcement learning with demonstrations.
\newblock In {\em 2018 IEEE International Conference on Robotics and Automation
  (ICRA)}, pages 6292--6299. IEEE, 2018.

\bibitem{peng2017deeploco}
Xue~Bin Peng, Glen Berseth, KangKang Yin, and Michiel Van De~Panne.
\newblock Deeploco: Dynamic locomotion skills using hierarchical deep
  reinforcement learning.
\newblock {\em ACM Transactions on Graphics (TOG)}, 36(4):41, 2017.

\bibitem{ross2011dagger}
St{\'e}phane Ross, Geoffrey Gordon, and Drew Bagnell.
\newblock A reduction of imitation learning and structured prediction to
  no-regret online learning.
\newblock In {\em Proceedings of the fourteenth international conference on
  artificial intelligence and statistics}, pages 627--635, 2011.

\bibitem{silver2012active}
David Silver, J~Andrew Bagnell, and Anthony Stentz.
\newblock Active learning from demonstration for robust autonomous navigation.
\newblock In {\em 2012 IEEE International Conference on Robotics and
  Automation}, pages 200--207. IEEE, 2012.

\bibitem{sun2017deeply}
Wen Sun, Arun Venkatraman, Geoffrey~J Gordon, Byron Boots, and J~Andrew
  Bagnell.
\newblock Deeply aggrevated: Differentiable imitation learning for sequential
  prediction.
\newblock In {\em Proceedings of the 34th International Conference on Machine
  Learning-Volume 70}, pages 3309--3318. JMLR. org, 2017.

\bibitem{sutton1999between}
Richard~S Sutton, Doina Precup, and Satinder Singh.
\newblock Between mdps and semi-mdps: A framework for temporal abstraction in
  reinforcement learning.
\newblock {\em Artificial intelligence}, 112(1-2):181--211, 1999.

\bibitem{todorov2012mujoco}
Emanuel Todorov, Tom Erez, and Yuval Tassa.
\newblock Mujoco: A physics engine for model-based control.
\newblock In {\em 2012 IEEE/RSJ International Conference on Intelligent Robots
  and Systems}, pages 5026--5033. IEEE, 2012.

\bibitem{vezhnevets2017feudal}
Alexander~Sasha Vezhnevets, Simon Osindero, Tom Schaul, Nicolas Heess, Max
  Jaderberg, David Silver, and Koray Kavukcuoglu.
\newblock Feudal networks for hierarchical reinforcement learning.
\newblock In {\em Proceedings of the 34th International Conference on Machine
  Learning-Volume 70}, pages 3540--3549. JMLR. org, 2017.

\end{thebibliography}
\end{document}